\documentclass[letterpaper, 10 pt, conference]{ieeeconf}  

\IEEEoverridecommandlockouts                              

\usepackage{graphicx} 
\usepackage{cite}
\usepackage{amsmath} 
\usepackage{amssymb}  
\usepackage{array}

\usepackage[colorinlistoftodos]{todonotes}

\title{\LARGE \bf
The role of haptic communication in dyadic collaborative object manipulation tasks
}

\author{Yiming Liu$^{1}$, Raz Leib$^{1}$, William Dudley$^{2,3}$, Ali Shafti$^{2}$, A. Aldo Faisal$^{2,3,4,*}$ and David W. Franklin$^{1,5,*}$
\thanks{*This work was enabled by the Imperial-TUM Joint Academy of Doctoral Studies (JADS), where YL was supported by the TUM International Graduate School of Science and Engineering (IGSSE) and WD was funded through the UKRI Centre for Doctoral Training in AI for Healthcare (EP/S023283/1). AS was supported by an EPSRC Network+ \emph{Human-Like Computing} Kick-start grant. AAF acknowledges a UKRI Turing AI Fellowship Grant (EP/V025449/1). The funders were not involved in the design or publication of this study. The authors declare no competing financial interests.}
\thanks{
$^{1}$Neuromuscular Diagnostics, Department of Sport and Health Sciences, Technical University of Munich, Munich, 80992 Germany.}
\thanks{
$^{2}$ Brain \& Behaviour Lab, Departments of Computing \& Bioengineering, Imperial College London, London, SW7 2AZ, United Kingdom.}
\thanks{$^{3}$ UKRI Centre in AI for Healthcare, Imperial College London, SW7 2AZ, London, UK.}
\thanks{$^{4}$ Institute of Artificial \& Human Intelligence (IAHI), University of Bayreuth, Germany}
\thanks{$^{5}$ Munich Institute of Robotics and Machine Intelligence (MIRMI), Technical University of Munich, Munich, 80992 Germany}
\thanks{$^{*}$ AAF and DWF are corresponding authors: {\tt\small \newline \indent{aldo.faisal@imperial.ac.uk} \newline \indent{david.franklin@tum.de}}}%
}

\begin{document}

\maketitle
\thispagestyle{empty}
\pagestyle{empty}

\begin{abstract}

Intuitive and efficient physical human-robot collaboration relies on the mutual observability of the human and the robot, i.e. the two entities being able to interpret each other's intentions and actions. This is remedied by a myriad of methods involving human sensing or intention decoding, as well as human-robot turn-taking and sequential task planning. However, the physical interaction establishes a rich channel of communication through forces, torques and haptics in general, which is often overlooked in industrial implementations of human-robot interaction. In this work, we investigate the role of haptics in human collaborative physical tasks, to identify how to integrate physical communication in human-robot teams. We present a task to balance a ball at a target position on a board either bimanually by one participant, or dyadically by two participants, with and without haptic information. The task requires that the two sides coordinate with each other, in real-time, to balance the ball at the target. We found that with training the completion time and number of velocity peaks of the ball decreased, and that participants gradually became consistent in their braking strategy. Moreover we found that the presence of haptic information improved the performance (decreased completion time) and led to an increase in overall cooperative movements. Overall, our results show that humans can better coordinate with one another when haptic feedback is available. These results also highlight the likely importance of haptic communication in human-robot physical interaction, both as a tool to infer human intentions and to make the robot behaviour interpretable to humans.

\end{abstract}

\section{INTRODUCTION}
The emergency of practical applications for collaborative robots has led to various platforms for safe human-robot interaction, resulting in a human in-the-loop system with unique challenges particularly in mutual observability between the human and the robot \cite{Ogenyi2021}. The lack of observability of the robot for the human can arise, e.g., from the use of control algorithms and therefore motions that are not interpretable for the human. From the robot's side, the lack of observabiltiy of the human can arise from, e.g., the stochasticity of human behaviour and challenges in human intent recognition. Crucially, both entities face issues in obtaining a theory of mind of each other, which can lead to lack of trust by the human, resulting in unsafe and unsuccessful interactions \cite{WouRobotTrust2020}.

\begin{figure}[tb]
    \centering
    \includegraphics[width=\columnwidth]{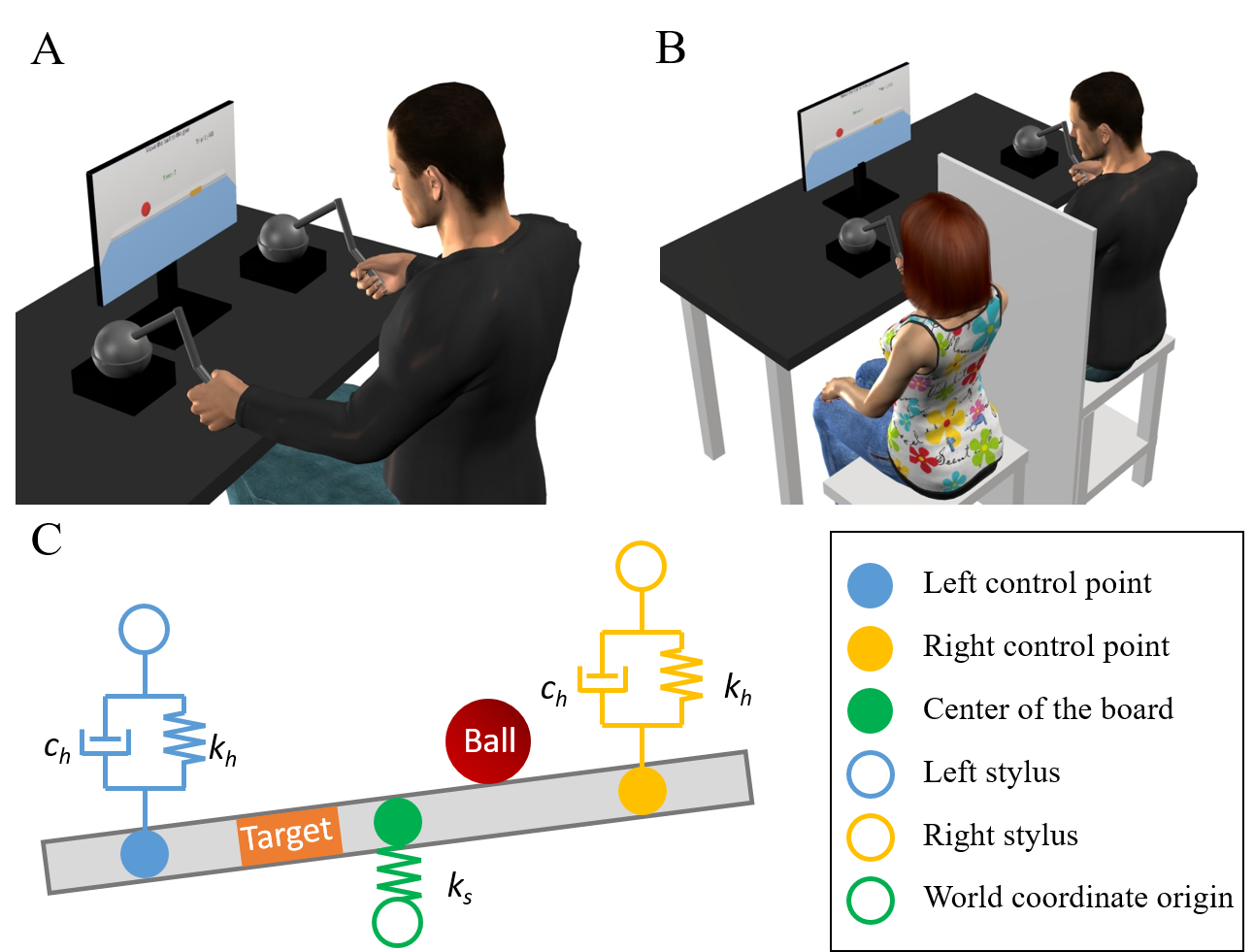}
    \caption{The experiment setup. A and B show the bimanual and dyadic experiment setup, respectively. The two participants in the dyadic experiments were separated by a curtain, such that they could not see the movement of their partners. The task is to control the board and balance the red ball within the orange target area. C, The dynamics of the virtual experiment model. The board was controlled by the participants via a spring-damper system and the center of the board was attached to the origin of the world coordinate by a spring.}
    \label{figuresetup}
\end{figure}

A common approach to solving issues of observability in human-robot collaboration, has been to rely on turn-taking approaches \cite{Ogenyi2021, MUKHERJEE2022102231}, and methods such as planning and scheduling to create a human-robot choreography \cite{PELLEGRINELLI20171}, to coordinate the two entities towards successful collaboration. While this has been applied successfully in industrial settings, it differs from how humans collaborate with each other in physical tasks\cite{EvrardHomotopy2009}. When it comes to physical interaction, we have access to a physical channel of communication through haptics \cite{reedPHRC2008,EvrardHomotopy2009,noohiHHC2016}, which can be leveraged to enhance mutual observability of the human-robot team. Using the physical channel, humans communicate through forces, effectively implementing the turn-taking approach at a micro-interaction level, through assigning leader and follower roles, and switching roles as needed to fulfil the task efficiently \cite{liRoleAdapt2015,messeriLF2020}.

We previously demonstrated the applicability of human in-the-loop reinforcement learning to learn personalised collaborative policies through direct, real-world and real-time interaction with a human partner, in a synchronous collaborative task \cite{shaftiHRC2020}. To extend our approach to physical interactions, we want to better understand the role of haptics in human-human and human-robot collaboration, and expose our reinforcement learning agent to the physical channel of communication. Here, we present an experimental setup to study human-human and human-robot haptic communication during physical collaborations (Fig. \ref{figuresetup}). We used two haptic devices (Phantom Touch; 3D SYSTEMS) to interface humans with a virtual reality setup involving balancing a ball at given target locations on a board. Our participants use the haptic devices bimanually, and in dyadic teams, in blocks of trials with and without haptic feedback, allowing us to study the role of haptics and dyadic coordination in the success of the collaboration. 

Many previous studies have examined dyadic human-human and human-robot physical interaction. When physical interaction takes place, haptic communication transfers valuable information via forces and torques sensed by the partners. This information can be used to infer intentions as well as to determine the role of each participant in the task, thus allows the dyad to better collaborate \cite{reedPHRC2008,EvrardHomotopy2009, losey2018review}. Humans interacting in a group can coordinate their movement by estimating the partner's goal \cite{takagi2017physically, takagi2019individuals}. When the two participants were coupled by a hard mechanics, they were more likely to develop into different roles and achieve a better performance, compared to when their interaction was soft \cite{takagi2018haptic, takai2021leaders}. Several models have been developed to allow robots to better collaborate with humans by inferring the human's intention from haptic information and adapting roles in the collaboration \cite{liRoleAdapt2015, EvrardHomotopy2009, noohiHHC2016}.

Much previous work studying human-human physical collaboration focused on simple tasks, such as moving a crank \cite{reedPHRC2008}, tracking a moving target \cite{takagi2017physically, takagi2018haptic, takagi2019individuals} or reaching movement \cite{takai2021leaders}. In this work, we studied the manipulation of a complex object with internal dynamics. The experiment was performed by individuals bimanually or by dyads either with or without haptic feedback. By comparing the performance in the two haptic conditions, we found that haptic feedback allowed the participants to coordinate better and achieve better performance.

\section{MATERIALS AND METHODS}

\subsection{Experimental apparatus}

Participants were asked to control the rotation and vertical position of a board in order to move and balance a ball within a target area in virtual reality (VR). The two ends of the board were controlled by two haptic devices, either bimanually by one participant, or dyadicly by two participants. These two configurations were examined both with and without haptic communication. When participants worked in dyadic teams, they were not allowed to talk and were separated by a curtain, such that they could not see their partners, see Fig.~\ref{figuresetup}. The VR environment was rendered by Chai3D \cite{conti2005chai}. Data in the VR environment, for example position of the ball, rotation of the board, and interaction forces were sampled at 1000 Hz. 

The positive directions of the X, Y and Z axes of the virtual environment were perpendicular to the screen pointing outward, to the right of the screen, and to the top of the screen, respectively. The board was 700 mm long and 50 mm wide. The radius of the ball was 25 mm. The 70 mm-wide target was placed at 150 mm to the left or right of the board. The center of the board was connected to the origin of the world coordinate by an invisible spring. Two control points were placed at 250 mm to the left and right of the center of the board. Participants held the styluses of two haptic devices, which were placed in front of the participants. The z-coordinates of the control points were controlled by the participants. The interaction forces between the board and the participants were computed with a spring-damper mechanism. Participants could control the position and angle of the board by moving the styluses of the haptic devices. The hand movement was restricted by the program to the vertical direction. The board could only be translated on the z axis and rotate around x axis, but it could not move on the x, y axes or rotate around y, z axes. The ball could only move along the long side of the board.

\subsection{Experimental paradigm}
At the beginning of each trial, the ball was fixed at the position opposite to the target area with respect to the center of the board. Participants were asked to tilt the board to get the center of the ball within the target area. Participants needed to keep the ball on the board. Once the ball fell off the edge of the board, the trial would be marked as failed and started over. The trial ended when the ball stayed inside the target area for 1.5 s. The ball was then shifted to the center of the current target. Then the target area switched to the other side of the board and the next trial started. Participants needed to keep the board horizontal, before the next trial started. Participants could see the completion time after each trial and were encouraged to finish the trials as fast as possible.

Each participant performed 12 blocks of experiments (18 blocks across a pair of participants), where each block consisted of 60 successful trials. In each block, the control of the board was either bimanual, i.e. one participant control the board using both hands, or dyadic, i.e. two participants each control one side of the board using their dominant hand. The two participants of each pair first alternately performed 4 bimanual blocks each, after that they cooperatively performed 6 dyadic blocks, and finally ended with another 2 bimanual blocks each. They could get haptic feedback in the z axis from the haptic devices in half of the blocks, while in the other half they could not. The haptic condition (on/off) was switched between each block, where the order of haptic conditions was counterbalanced across all participants/pairs. For each condition (bimanual/dyadic, haptic on/off), each participant had 180 trials in total.

\subsection{Virtual system model}
Participants could apply forces to the board at the two control points. The forces were generated according to Equation \ref{eqn:force}.

\begin{equation}
\label{eqn:force}
F_n=k_h~(z_{stylus,n}-z_{point,n})+c_h~(\Dot{z}_{stylus,n}-\Dot{z}_{point,n})
\end{equation}

where n is either left or right, $F_{\mbox{\scriptsize{n}}}$ is the generated force, $z_{\mbox{\scriptsize{stylus}}}$ and $z_{\mbox{\scriptsize{point}}}$ are the z coordinates of the stylus and the control point, respectively.  
The motion of the board and the ball were simulated according to Equation \ref{eqn:boardpos}, \ref{eqn:boardangle} and \ref{eqn:ballpos}.

\begin{gather}
F_s=-k_s~z_{board} \label{eqn:spring}\\
F_{left}+F_{right}-M~g-m~g~\cos^{2}\theta+F_s=M\Ddot{z}_{board} \label{eqn:boardpos}\\
(F_{right}-F_{left})~l~\cos\theta-m~g~p_{ball}\cos\theta=I~\Ddot{\theta}  \label{eqn:boardangle}\\
m~\Ddot{p}_{ball} = m~g~\sin \theta \label{eqn:ballpos}
\end{gather}

In these equations, $F_{\mbox{\scriptsize{s}}}$ is the force generated by the spring connecting the board to the origin of the world coordinate. $z_{\mbox{\scriptsize{board}}}$ is the z coordinate of the center of the board. $\theta$ is the angle of the board around the x axis. Positive values means the board is rotated counter clockwise. $p_{\mbox{\scriptsize{ball}}}$ is the relative position of the ball with respect of the center of the board along the ling side of the board, with positive values meaning the ball is to the right of the board. The values and meanings of the coefficients in the equations above are summarized in Table \ref{tbl:coefficients}.

\newcolumntype{C}[1]{>{\centering\let\newline\\\arraybackslash\hspace{0pt}}m{#1}}
\begin{table}[h!]
\centering
\begin{tabular}{|C{0.9cm}|C{4cm}|C{0.9cm}|C{0.9cm}|} 
 \hline
 Symbol & Coefficient & Unit & Value \\ [0.5ex] 
 \hline
 $M$ & Weight of the board & $kg$ & 0.01 \\ 
 \hline
 $m$ & Weight of the ball & $kg$ & 0.05 \\
 \hline
 $g$ & Gravity acceleration & $m/s^2$ & 9.81 \\
 \hline
 $k_h$ & Force input spring stiffness & $N/s$ & 200 \\
 \hline
 $c_h$ & Force input damper coefficient & $Ns/m$ & 2 \\
 \hline
 $k_s$ & Stiffness of the spring connected to the origin & $N/m$ & 140 \\
 \hline
 $l$ & distance between the control and the center of the board & $m$ & 0.25 \\
 \hline
 $I$ & Moment of inertia of the board & $kg \cdot m^2$ & 0.0004\\[1ex] 
 \hline
\end{tabular}
\caption{Coefficients of the virtual system model.}
\label{tbl:coefficients}
\end{table}

For the blocks with haptic feedback, $-F_{left}$ and $-F_{right}$ were generated by the left and right haptic devices, respectively. No forces in the z axis were generated in blocks without haptic feedback.

\subsection{Data analysis}
After data collection, the force and kinematic data were low-pass filtered with a tenth-order, zero-phase-lag Butterworth filter with 20 Hz cutoff frequency to remove any high frequency noise from data recording. 
\subsubsection{Completion time}

Completion time was defined as the time from the beginning of each trial, to the beginning of the time when the ball stayed in the target area for 1.5s. 

\subsubsection{Participants' Movement classification}
At each time step, the movement of the participants were classified into 4 types based on the velocity in the z axis of the left and right styluses. Note that if the magnitude of velocity is lower than 0.003 m/s, the stylus is considered as not moving. 
\begin{itemize}
  \item Cooperative movement: the two sides were moving in opposite directions.
  \item Competitive movement: the two sides were moving in the same direction.
  \item Single movement: one side was moving while the other side was not. 
  \item Still: Both sides were not moving. 
\end{itemize}

\subsubsection{Board movement segmentation and delay}
Each trial was divided into multiple segments based on the angular velocity of the board. The angular velocity was first low-pass filtered with a tenth-order, zero-phase-lag Butterworth filter with 5 Hz cutoff frequency, in order to make the segmentation less sensitive to noise. If the filtered angular velocity was higher than 1 degree/s for at least 100 ms and the angle changed by at least 1 degree, this period was marked as a board movement segment. 

We calculated the delay between the left and right hands (or left and right participants) for all cooperative movement segments. A segment was marked as cooperative if the styluses moved in opposite directions within the segment and one side moved no more than nine times as much as the other side. For all cooperative segments, the delay was calculated as the difference in time when each side reached a percentage of the peak velocity of the slower side in that segment. This was done for 4 different percentages (10\%, 20\%, 30\% and 40\%) to ensure the result is robust to the chosen threshold. 

\subsubsection{Number of peaks in the ball absolute velocity profile}
The number of peaks (NP) in the absolute velocity profile of each trial was calculated according to Equation \ref{eqn:smoothness}. 

\begin{equation}
\label{eqn:smoothness}
NP=\Big | \Big \{ v_{ball}(t), \frac{dv_{ball}(t)}{dt}=0, \frac{d^2v_{ball}(t)}{dt^2} \Big \} \Big |
\end{equation}

where $|\cdot |$ represents the cardinality of a set, $v_ball$ is the absolute velocity of the ball with respect to the center of the board. To make the calculation less sensitive to noise, $v_ball$ was first low-pass filtered with a tenth-order, zero-phase-lag Butterworth filter with 5 Hz cutoff frequency.

\subsubsection{Strategy}
In order to balance the ball within the target range as fast as possible, the participants need to develop a control strategy. Here we focus on the first braking movement. The braking movement was defined as the board movement segments that generated an acceleration on the ball in the opposite direction to the target. We considered the onset of the first braking movement after the initial movement towards the target. We hypothesized that after the initial movement, participants would wait for the ball to reach a specific position and velocity before performing the first braking movement. The strategy was represented by two values: a) The distance between the ball and the midpoint of the target, b) The velocity of the ball in the direction of the target. Both at the onset of the first braking movement. 

We divided the 180 trials for each participant/dyad under each haptic condition into 6 groups with 30 trials each, in the order of completion. The distribution of the strategies of each group were represented by an ellipse. The center of the ellipse is the mean of the strategies. We used Principal Component Analysis on each group and set the rotation of the ellipse to the direction of the first principal component. The width and height of the ellipse are calculated by the 90th percentile of the range of variation of the data in the direction of the first and second principal component.

\section{RESULTS}
Seven right-handed participants and one left-handed participant \cite{oldfield1971assessment} (23-28 years of age, 2 women) took part in this study. All participants were neurologically healthy. They were naive to the purpose of this study and provided written informed consent before participation. The study was approved by the institutional ethics committee at the Technical University of Munich.

\subsection{Learning effect}
\begin{figure}[tb]
    \centering
    \includegraphics[width=\columnwidth]{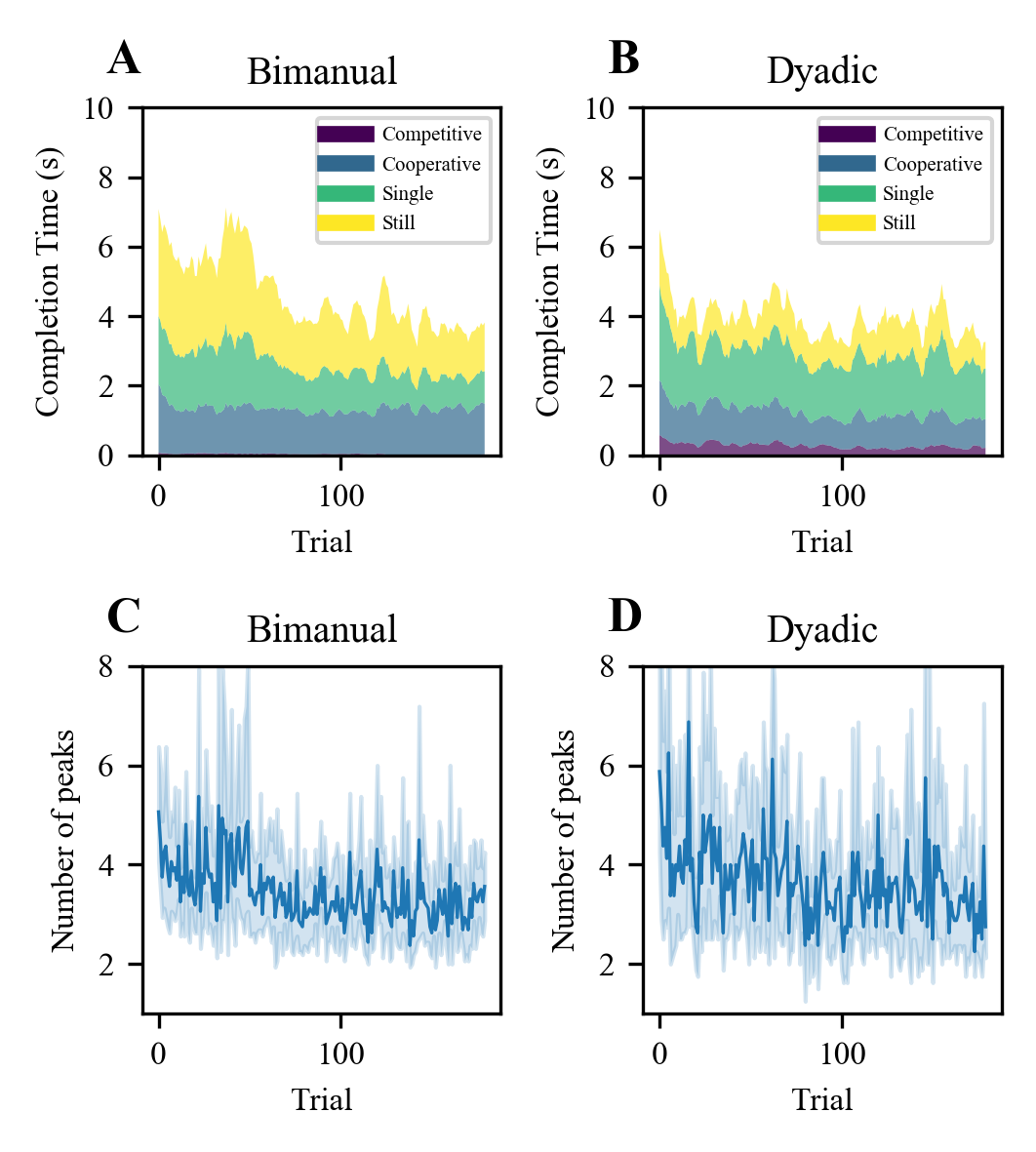}
    \caption{The learning effect. A and B show the completion time and share of each type of movement as a function of trials of the bimanual and dyadic experiment, respectively. C and D show the number of peaks in the ball velocity profile of the bimanual and dyadic experiment, respectively. All plots are averaged across all participants/dyads and both with or without haptic feedback conditions}
    \label{learning_effect}
\end{figure}

Participants' performance improved over the entire experiment. The mean and standard deviation of the completion time decreased from 5.71$\pm$4.48 s (mean$\pm$std) in the first 30 trials to 3.61$\pm$3.08 s in the last 30 trials for the bimanual experiment and from 4.33$\pm$2.92 s in the first 30 trials to 3.62$\pm$2.23 s in the last 30 trials (Fig. \ref{learning_effect}A, B) for the dyadic experiment. The colors denote the time of the four types of movements. Comparing the bimanual and dyadic experiments, it is clear that the share of cooperative movements and still time was higher in bimanual experiments, while dyadic experiments had more single movements and competitive movements. As expected there was almost no competitive movements in the bimanual experiments. 

Another aspect of the learning effect is the number of peaks in the ball absolute velocity profile. Number of peaks decreased from 3.92$\pm$2.42 (mean$\pm$std) in the first 30 trials to 3.17$\pm$1.85 in the last 30 trials for the bimanual experiment and from 4.27$\pm$2.91 in the first 30 trials to 3.28$\pm$1.95 in the last 30 trials for the dyadic experiment.

\subsection{The role of haptic feedback}
\begin{figure}[tb]
    \centering
    \includegraphics[width=\columnwidth]{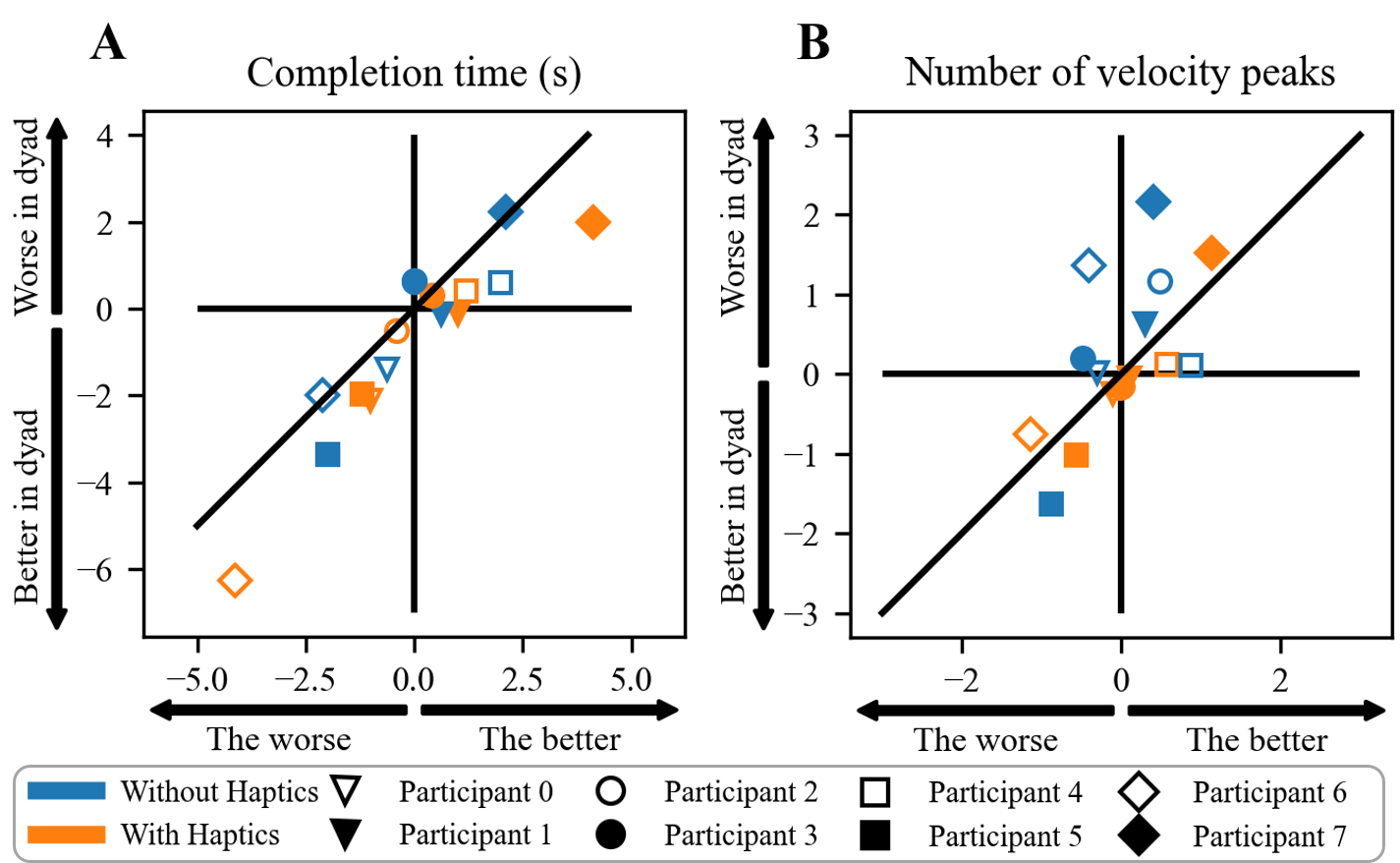}
    \caption{Comparison between the performance of dyadic and bimanual experiments. A and B show the performance in completion time and trajectory smoothness, respectively. The x and y coordinates show the relative performance of each participant with respect to their partners and their dyad, respectively.}
    \label{DyadImprovement}
\end{figure}

For each dyad, we compared individual performance and dyadic performance with or without haptic feedback, to see if performance improved in dyad. In Figure \ref{DyadImprovement}, the x and y coordinates were calculated as follows: $x = \frac{1}{2}(v_{partner} - v_{self})$, $y = v_{dyad} - v_{self} $ where v is either completion time or number of peaks, the footnote of $self$ and $partner$ denote the bimanual performance of each participant and their partners, respectively, and $dyad$ denote the performance of each dyad. A positive x coordinate means that the participant had a better performance than their partner, and a positive y coordinate means that the participant's performance deteriorated compared to their bimanual performance. If the dyadic performance is better than the average performance of the two participants, the data points should be below the diagonal line. When there was haptic feedback, all dyads had a shorter completion time as compared to the average individual completion time, while half of the dyads took a longer time to complete the task when there was no haptic feedback (Fig. \ref{DyadImprovement}A). A similar result could be found in the smoothness of ball trajectory. Three out of four dyads had a smoother trajectory when there was haptic feedback, while only one dyad had a smoother trajectory when there was no haptic feedback (Fig. \ref{DyadImprovement}B). 

\begin{figure}[tb]
    \centering
    \includegraphics[width=\columnwidth]{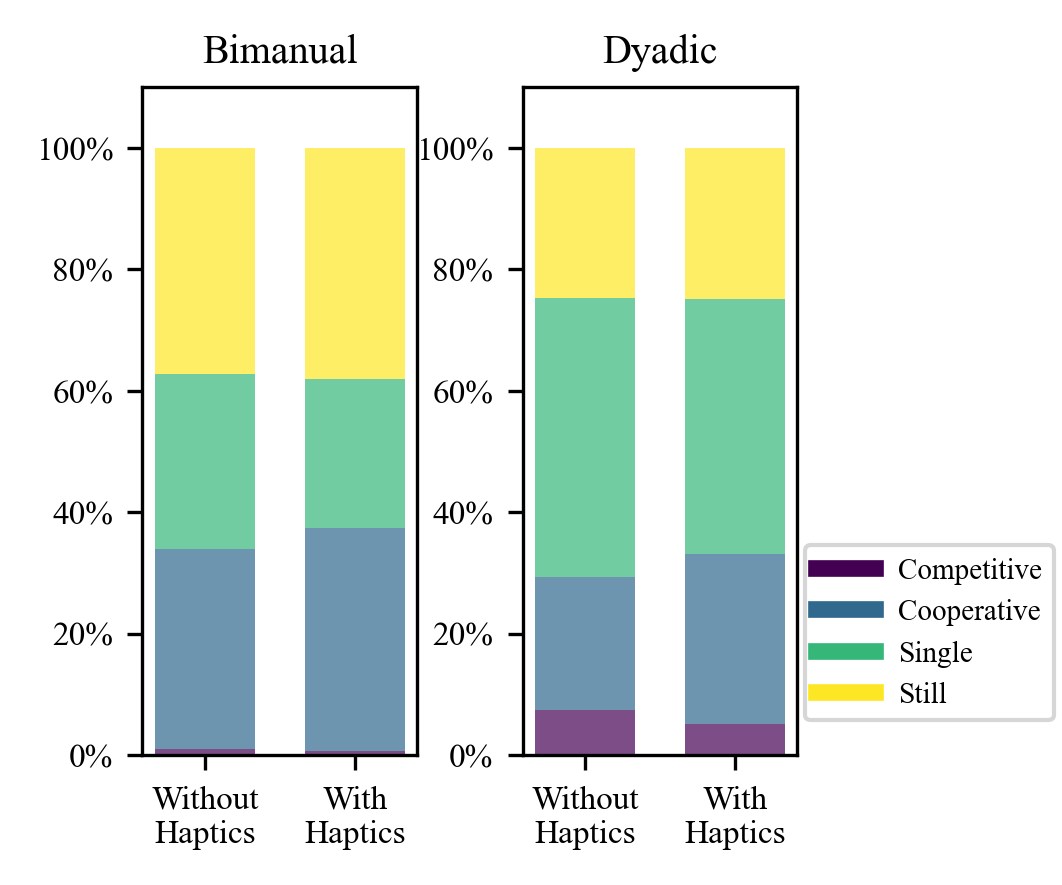}
    \caption{Comparison of the movement classification between the two haptic conditions of the dyadic and bimanual experiments. The share of each movement class are shown in different colors.}
    \label{MovementClassification}
\end{figure}

We also investigated whether the presence of haptic feedback influences the way that two hands or two participants cooperate. We compared the movement classification between the two haptic conditions both in the bimanual and dyadic configurations (Fig. \ref{MovementClassification}). When there was haptic feedback, the share of cooperative movement increased and the share of single movement decreased for both bimanual and dyadic experiments. The share of competitive movement also decreased in the dyadic experiments.

\subsection{Convergence of Strategy}
\begin{figure}[htp]
    \centering
    \includegraphics[width=\columnwidth]{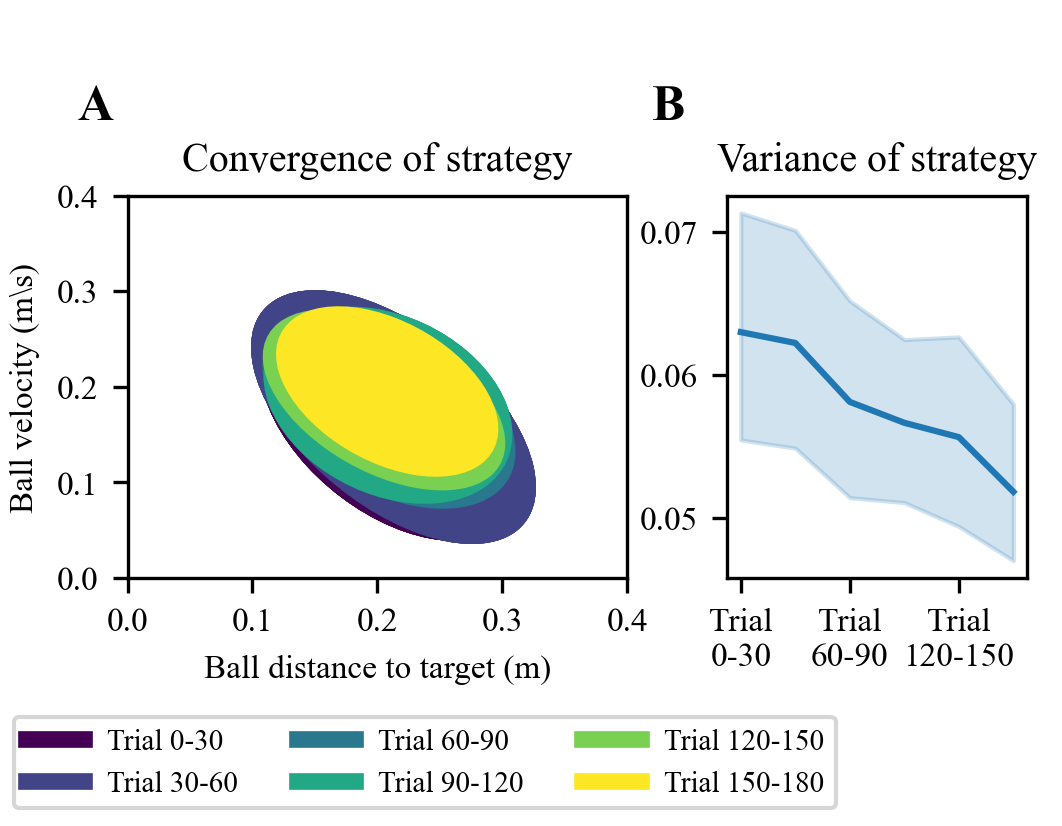}
    \caption{Convergence of strategy. A, each ellipse represent the ball velocity and distance to the target of a group of 30 trials for all participants and dyads. B, the variance of the ball velocity and distance to the target decreased over trials. The shaded area indicates the 95\% confidence interval of the mean. }
    \label{Convergence}
\end{figure}

To stabilize the ball within the target area in the shortest possible time, participants needed to start braking when the ball is at a specific velocity and distance to the target. We divided the 180 trials of each condition into 6 groups of 30 trials according to the experimental order. For each group of trials, we used ellipses to represent the range of ball velocity and distance to the target (See MATERIALS AND METHODS). The size of the ellipse became smaller over time, which indicate that participants' strategy converged to a subset (Fig. \ref{Convergence}A). Moreover the variance of the strategy decreased over trials (Fig. \ref{Convergence}B), which shows that the strategies became more stable with more practice. This variance was calculated for each participant/dyad.

\subsection{Delay}
\begin{figure}[tb]
    \centering
    \includegraphics[width=6.5cm]{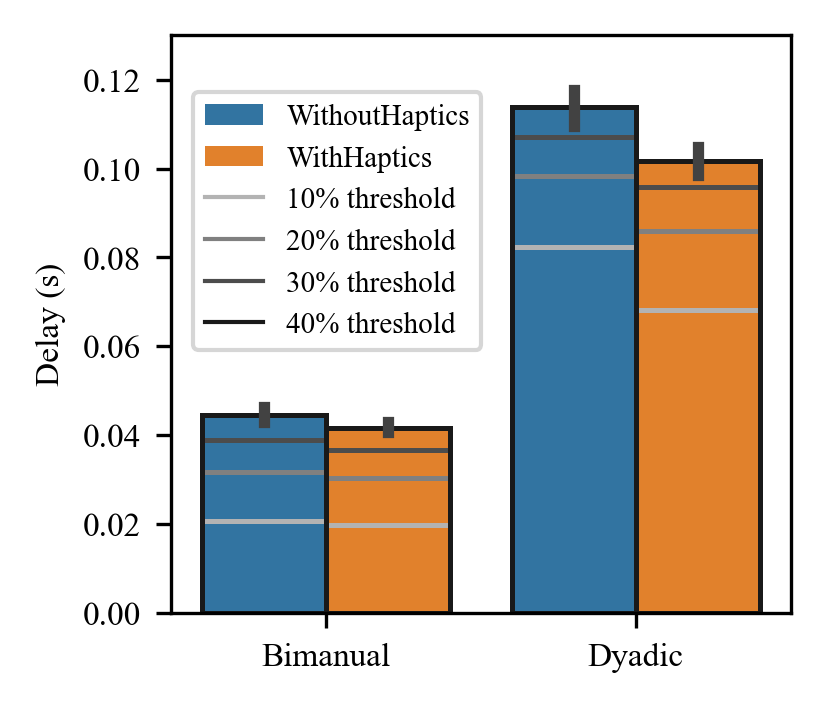}
    \caption{Absolute delay of each condition during the cooperative movements. The means calculated by each threshold are marked by different grey levels. The error bars show 95\% confidence interval of the mean.}
    \label{Delay}
\end{figure}

The delay between the the handles was calculated during all of the cooperative movements during the experiments, using four different thresholds. Although the amount of delay varied depending on the threshold chosen (increasing with the threshold percentage), the relative differences across conditions were consistent for all four thresholds (Fig \ref{Delay}). The delay was always higher in dyadic experiments compared with bimanual experiments, with a delay around 100 ms in the dyadic experiments and only about 40 ms in the bimanual experiments. Moreover, the delay in dyadic experiments significantly decreased when there was haptic feedback (95\% confidence intervals do not overlap, Fig \ref{Delay}), whereas there was little affect on the delay in the bimanual experiments with the presence of haptic feedback.

\addtolength{\textheight}{-5cm}   

\section{DISCUSSION}
In this work, we investigated the role of haptic feedback in bimanual and dyadic coordination of humans, to inform the use of the physical channel of communication in physical human-robot collaboration. Our result shows that participants moved with shorter delay, better coordination and achieved better performance when they can receive haptic feedback. This is consistent with several previous studies. Takagi and colleagues \cite{takagi2018haptic} showed that the performance of the worse partner improved more when the interaction was stiffer, at the cost of the better partner having to exert more effort. Their model explained the process by assuming that the partner's target was inferred through haptics and tuned by the coupling stiffness. Takai and colleagues \cite{takai2021leaders} showed that participants were only able to co-adapt and develop into different roles when there was haptic feedback. Our work extends these findings, suggesting that haptic communication is an important channel carrying information that allows humans to infer the intention of their partners. This becomes even more critical in our more complex task, where the two partners must coordinate their actions in order to successfully complete the task due to the internal degrees of freedom. These effects were seen despite the fact that the haptic feedback in our experiment was limited. First, the center of the board was not fixed by a hinge, but was connected by a spring resulting in small forces on each hand. Secondly our haptic devices could only generate forces, but not torques. Nonetheless, we saw a clear effect of haptic communication in the dyadic coordination task, both in terms of improved performance and a reduction in the delay between the two humans.

Our work highlights the expected importance of haptic communication during physical interactions between humans and robots while performing collaborative tasks. Several studies have used haptic information to understand human intention and enable robots to adjust accordingly\cite{liRoleAdapt2015, EvrardHomotopy2009, noohiHHC2016}. However, haptic information may also have a large influence on the coordination between the robot and human. Not only should haptic information allow robots to infer more precise inference of human intention, but it also allows them to generate appropriate interaction forces and torques to signal their strategy to the human. This may make up for the gap of mutual observability and lack of theory of mind between collaborating humans and robots previously identified as a challenge in real-time human-robot collaboration\cite{shaftiHRC2020}.




\bibliographystyle{IEEEtran}
\bibliography{bibtex/IEEEabrv, bibtex/mybibfile}

\begin{thebibliography}{10}
\providecommand{\url}[1]{#1}
\csname url@samestyle\endcsname
\providecommand{\newblock}{\relax}
\providecommand{\bibinfo}[2]{#2}
\providecommand{\BIBentrySTDinterwordspacing}{\spaceskip=0pt\relax}
\providecommand{\BIBentryALTinterwordstretchfactor}{4}
\providecommand{\BIBentryALTinterwordspacing}{\spaceskip=\fontdimen2\font plus
\BIBentryALTinterwordstretchfactor\fontdimen3\font minus
  \fontdimen4\font\relax}
\providecommand{\BIBforeignlanguage}[2]{{%
\expandafter\ifx\csname l@#1\endcsname\relax
\typeout{** WARNING: IEEEtran.bst: No hyphenation pattern has been}%
\typeout{** loaded for the language `#1'. Using the pattern for}%
\typeout{** the default language instead.}%
\else
\language=\csname l@#1\endcsname
\fi
#2}}
\providecommand{\BIBdecl}{\relax}
\BIBdecl

\bibitem{Ogenyi2021}
U.~E. Ogenyi, J.~Liu, C.~Yang, Z.~Ju, and H.~Liu, ``Physical human-robot
  collaboration: Robotic systems, learning methods, collaborative strategies,
  sensors, and actuators,'' \emph{IEEE Transactions on Cybernetics}, vol.~51,
  pp. 1888--1901, 4 2021.

\bibitem{WouRobotTrust2020}
W.~Mou, M.~Ruocco, D.~Zanatto, and A.~Cangelosi, ``When would you trust a
  robot? a study on trust and theory of mind in human-robot interactions,'' in
  \emph{2020 29th IEEE International Conference on Robot and Human Interactive
  Communication (RO-MAN)}, 2020, pp. 956--962.

\bibitem{MUKHERJEE2022102231}
\BIBentryALTinterwordspacing
D.~Mukherjee, K.~Gupta, L.~H. Chang, and H.~Najjaran, ``A survey of robot
  learning strategies for human-robot collaboration in industrial settings,''
  \emph{Robotics and Computer-Integrated Manufacturing}, vol.~73, p. 102231,
  2022. [Online]. Available:
  \url{https://www.sciencedirect.com/science/article/pii/S0736584521001137}
\BIBentrySTDinterwordspacing

\bibitem{PELLEGRINELLI20171}
\BIBentryALTinterwordspacing
S.~Pellegrinelli, A.~Orlandini, N.~Pedrocchi, A.~Umbrico, and T.~Tolio,
  ``Motion planning and scheduling for human and industrial-robot
  collaboration,'' \emph{CIRP Annals}, vol.~66, no.~1, pp. 1--4, 2017.
  [Online]. Available:
  \url{https://www.sciencedirect.com/science/article/pii/S0007850617300951}
\BIBentrySTDinterwordspacing

\bibitem{EvrardHomotopy2009}
P.~Evrard and A.~Kheddar, ``Homotopy switching model for dyad haptic
  interaction in physical collaborative tasks,'' in \emph{World Haptics 2009 -
  Third Joint EuroHaptics conference and Symposium on Haptic Interfaces for
  Virtual Environment and Teleoperator Systems}, 2009, pp. 45--50.

\bibitem{reedPHRC2008}
K.~B. Reed and M.~A. Peshkin, ``Physical collaboration of human-human and
  human-robot teams,'' \emph{IEEE Transactions on Haptics}, vol.~1, no.~2, pp.
  108--120, 2008.

\bibitem{noohiHHC2016}
E.~Noohi, M.~Žefran, and J.~L. Patton, ``A model for human–human
  collaborative object manipulation and its application to human–robot
  interaction,'' \emph{IEEE Transactions on Robotics}, vol.~32, no.~4, pp.
  880--896, 2016.

\bibitem{liRoleAdapt2015}
Y.~Li, K.~P. Tee, W.~L. Chan, R.~Yan, Y.~Chua, and D.~K. Limbu, ``Continuous
  role adaptation for human–robot shared control,'' \emph{IEEE Transactions
  on Robotics}, vol.~31, no.~3, pp. 672--681, 2015.

\bibitem{messeriLF2020}
C.~Messeri, A.~M. Zanchettin, P.~Rocco, E.~Gianotti, A.~Chirico, S.~Magoni, and
  A.~Gaggioli, ``On the effects of leader-follower roles in dyadic human-robot
  synchronisation,'' \emph{IEEE Transactions on Cognitive and Developmental
  Systems}, pp. 1--1, 2020.

\bibitem{shaftiHRC2020}
A.~Shafti, J.~Tjomsland, W.~Dudley, and A.~A. Faisal, ``Real-world human-robot
  collaborative reinforcement learning,'' in \emph{2020 IEEE/RSJ International
  Conference on Intelligent Robots and Systems (IROS)}, 2020, pp.
  11\,161--11\,166.

\bibitem{losey2018review}
D.~P. Losey, C.~G. McDonald, E.~Battaglia, and M.~K. O'Malley, ``A review of
  intent detection, arbitration, and communication aspects of shared control
  for physical human--robot interaction,'' \emph{Applied Mechanics Reviews},
  vol.~70, no.~1, 2018.

\bibitem{takagi2017physically}
A.~Takagi, G.~Ganesh, T.~Yoshioka, M.~Kawato, and E.~Burdet, ``Physically
  interacting individuals estimate the partner’s goal to enhance their
  movements,'' \emph{Nature Human Behaviour}, vol.~1, no.~3, pp. 1--6, 2017.

\bibitem{takagi2019individuals}
A.~Takagi, M.~Hirashima, D.~Nozaki, and E.~Burdet, ``Individuals physically
  interacting in a group rapidly coordinate their movement by estimating the
  collective goal,'' \emph{Elife}, vol.~8, p. e41328, 2019.

\bibitem{takagi2018haptic}
A.~Takagi, F.~Usai, G.~Ganesh, V.~Sanguineti, and E.~Burdet, ``Haptic
  communication between humans is tuned by the hard or soft mechanics of
  interaction,'' \emph{PLoS computational biology}, vol.~14, no.~3, p.
  e1005971, 2018.

\bibitem{takai2021leaders}
A.~Takai, Q.~Fu, Y.~Doibata, G.~Lisi, T.~Tsuchiya, K.~Mojtahedi, T.~Yoshioka,
  M.~Kawato, J.~Morimoto, and M.~Santello, ``Leaders are made: Learning
  acquisition of consistent leader-follower relationships depends on implicit
  haptic interactions.'' \emph{bioRxiv}, 2021.

\bibitem{conti2005chai}
F.~Conti, F.~Barbagli, D.~Morris, and C.~Sewell, ``Chai 3d: An open-source
  library for the rapid development of haptic scenes,'' \emph{IEEE World
  Haptics}, vol.~38, no.~1, pp. 21--29, 2005.

\bibitem{oldfield1971assessment}
R.~C. Oldfield, ``The assessment and analysis of handedness: the edinburgh
  inventory,'' \emph{Neuropsychologia}, vol.~9, no.~1, pp. 97--113, 1971.

\end{thebibliography}

\end{document}